\documentclass[10pt,twocolumn,letterpaper]{article}

\usepackage{iccv}
\usepackage{times,graphicx,amsmath,amssymb,balance,url}
\usepackage{booktabs,multirow,xspace,xfrac}
\usepackage[bf,small]{caption}
\usepackage[]{verbatim}
\usepackage[noblocks]{authblk}

\usepackage[breaklinks=true,bookmarks=false]{hyperref}

\iccvfinalcopy %

\def\iccvPaperID{****} %

\ificcvfinal\fi

\def\Vec#1{{\boldsymbol{#1}}}
\def\Mat#1{{\boldsymbol{#1}}}

\graphicspath{{./}{./figures/}}

\newcommand{\BlackBox}{\rule{1.5ex}{1.5ex}}  %
\newenvironment{proof}{\par\noindent{\bf Proof.\ }}{\hfill\BlackBox\\[2mm]}
\newtheorem{theorem}{Theorem}
\newtheorem{definition}[theorem]{Definition}

\def\grassmann{Grassmann\xspace}
\def\grassmannian{Grassmannian\xspace}
\newcommand{\tr}{\mathop{\rm  Tr}\nolimits}
\renewcommand{\Mat}[1]{\mathbf{#1}}

\begin{document}

\title
  {
  Dictionary Learning and Sparse Coding on Grassmann Manifolds: An Extrinsic Solution\thanks{Appearing in
      Proc. 14th International Conference on Computer Vision (ICCV), Dec.\ 2013, Sydney Australia.}
  }

\author[1,2]{Mehrtash Harandi}
\author[1,3]{Conrad Sanderson}
\author[4]{Chunhua Shen}
\author[5]{Brian C. Lovell}

\affil[1]{NICTA, Australia}
\affil[2]{Australian National University, Australia}
\affil[3]{Queensland University of Technology, Australia}
\affil[4]{University of Adelaide, Australia}
\affil[5]{University of Queensland, Australia}
\maketitle

\begin{abstract}
\vspace{-1ex}

\noindent
Recent advances in computer vision and machine learning suggest that a wide range of problems
can be addressed more appropriately by considering \mbox{non-Euclidean} geometry.
In this paper we explore sparse dictionary learning over the space of linear subspaces,
which form Riemannian structures known as \grassmann manifolds.
To this end, we propose to embed \grassmann manifolds into the space of symmetric matrices by an isometric mapping,
which enables us to devise a closed-form solution for updating a Grassmann dictionary, atom by atom.
Furthermore, to handle non-linearity in data, we propose a kernelised version of the dictionary learning algorithm.
Experiments on several classification tasks
(face recognition, action recognition, dynamic texture classification)
show that the proposed approach achieves considerable improvements in discrimination accuracy,
in comparison to state-of-the-art methods such as
kernelised Affine Hull Method and graph-embedding \grassmann discriminant analysis.
\end{abstract}

\vspace{-2ex}
\section{Introduction}
\label{sec:intro}

Linear subspaces of $\mathbb{R}^d$ can be considered as the core of
many inference algorithms in computer vision and machine learning.
For example, several state-of-the-art methods for matching videos or image sets
model given data by subspaces~\cite{HAMM2008_ICML,Harandi_CVPR_2011,Turaga_PAMI_2011}.
Auto regressive and moving average models, which are typically
employed to model dynamics in spatio-temporal processing,
can also be expressed as linear subspaces~\cite{Turaga_PAMI_2011}.
More applications of linear subspaces in computer vision include, but
are not limited to, chromatic noise filtering~\cite{Subbarao_2009_IJCV},
biometrics~\cite{Sanderson_AVSS12}, and
domain adaptation~\cite{DA_ICCV11}.

Despite their wide applications and appealing properties (\eg, the set of all reflectance functions produced by
Lambertian objects lies in a linear subspace), subspaces lie on a special type of Riemannian manifold,
namely the \grassmann manifold, which makes their analysis very challenging.
This paper tackles and provides efficient solutions to the following
two fundamental problems for learning on \grassmann manifolds:
\begin{enumerate}
\item
\itemsep -.11cm
    \emph{Sparse coding.}
Given a signal $\Mat{X}$ and a dictionary $\mathbb{D} = \{\Mat{D}_i\}_{i=1}^N$ with $N$ elements (also known as atoms),
where $\Mat{X}$ and $\Mat{D}_i$ are linear subspaces,
how $\Mat{X}$ can be approximated by a combination of a ``few'' atoms in~$\mathbb{D}$~?

\item {\emph{Dictionary learning.}
Given a set of measurements $\{\Mat{X}_i\}_{i=1}^m$,
how can a dictionary $\mathbb{D} = \{\Mat{D}_i\}_{i=1}^N$
be learned to represent $\{\Mat{X}_i\}_{i=1}^m$ sparsely~?
}
\end{enumerate}

Our main motivation here is to develop new methods for analysing video
data and image sets.
This is inspired by the success of sparse signal modelling that
suggests natural signals like images (and hence video and
image sets as our concern here) can be efficiently
approximated by superposition of atoms of a dictionary,
where the coefficients of superposition are usually sparse (\ie, most coefficients are zero).
    We generalise the traditional sparse coding, which operates on
    vectors, to sparse coding on subspaces.
    Sparse encoding with the dictionary of subspaces can then be seamlessly
    used for categorising video data.
    Before we present our main results, we want to highlight that
    the proposed algorithms outperform state-of-the-art methods on
    various recognition tasks and in particular has achieved the
    {\em highest} reported accuracy  in classifying dynamic textures.

{\bf{Related work.}}
While significant steps have been taken to develop the theory of the sparse coding and dictionary learning in Euclidean spaces,
similar problems on non-Euclidean geometry have received comparatively little attention~\cite{Harandi_ECCV12,Vemuri_MICCAI_2012}.
To our best knowledge, among a handful of solutions devised on Riemannian manifolds, none is specialised for
the \grassmann manifolds which motivates our study.

In \cite{Harandi_ECCV12}, the authors addressed sparse coding and dictionary learning for the Riemannian structure of Symmetric Positive Definite (SPD) matrices or tensors. The solution was obtained by embedding the SPD manifold into Reproducing Kernel Hilbert Space (RKHS) using a Riemannian kernel.
Another approach to learning a Riemannian dictionary is by exploiting
the tangent bundle of the manifold, as
for example in~\cite{Vemuri_MICCAI_2012} for the manifold of probability distributions. Since the
sparse coding has a trivial solution in this approach, an affine
constraint has to be added to the problem~\cite{Vemuri_MICCAI_2012}.
While having an affine constraint along with sparse coding is welcome in specific tasks
(\eg, clustering~\cite{Cetingul:CVPR:2009}),
in general, the resulting formulation is restrictive and no longer addresses the original problem.
Also, working in successive tangent spaces, though common,
values only a first-order approximation to the manifold at each step.
Furthermore, switching back and forth to the tangent spaces of a \grassmann manifold
(as required by this formulation) can be computationally very demanding for the
problems that we are interested in (\eg, video analysis).
This in turns makes the applicability of such school of thought
limited for the \grassmann manifolds arising in vision tasks.

{\bf Contributions}.
In light of the above discussion, in this paper, we introduce an extrinsic method for learning a \grassmann dictionary.
To this end, we propose to embed \grassmann manifolds into the space of symmetric matrices
by a diffeomorphism that preserves \grassmann projection distance
(a special class of distances on \grassmann manifolds).
We show how sparse coding can be accomplished in the induced space
and devise a closed-form solution for updating a \grassmann dictionary atom by atom.
Furthermore, in order to accommodate non-linearity in data, we propose a kernelised version of our dictionary learning algorithm.
Our contributions are therefore three-fold:
\begin{itemize}
        \itemsep -.1cm
\item[1.]
    We propose an extrinsic dictionary learning algorithm for data points on
    \grassmann manifolds by embedding the manifolds into the space of
    symmetric matrices.
\item[2.]
    We derive a kernelised version of the dictionary learning
    algorithm which can address the non-linearity in data.
\item[3.]
    We apply the proposed \grassmannian dictionary learning methods to
    several computer vision tasks where the data are videos or image sets.
    Our proposed algorithms outperform  state-of-the-art methods
    on a wide range of classification tasks,
    including face recognition from image sets, action recognition and dynamic texture classification.
\end{itemize}

\section{Background}
\label{sec:problem_def}

        Before presenting our algorithms, we review some concepts of
        Riemannian geometry of \grassmann manifolds, which
        provide the grounding for the proposed algorithms.
        Details on \grassmann manifolds and related topics can be
        found in~\cite{Absil:2008}.

\textbf{Geometry of \grassmann manifolds}.
The space of $d \times p$ ($0<p<d$) matrices with orthonormal columns
is a Riemannian manifold known as a Stiefel manifold {$\mathrm{St}(p,d)$},
\ie, {$\mathrm{St}(p,d) \triangleq \{\Mat{X} \in \mathbb{R}^{d \times p}: {\bf X}^ T {\bf X} = {\bf I}_p \} $}
where $ {\bf I}_p $ denotes the identity matrix of size $ p \times p $~\cite{Absil:2008}.
\grassmann manifold {$\mathcal{G}(p,d)$} can be defined
as a quotient manifold of {$\mathrm{St}(p,d)$} with the equivalence relation $\sim$ being:
{
{$\Mat{X}_1 \sim \Mat{X}_2$} if and only if {$\operatorname{Span}\left(\Mat{X}_1 \right) = \operatorname{Span}\left(\Mat{X}_2 \right)$}},
where {$\operatorname{Span}(\Mat{X})$} denotes the subspace spanned by columns of {$\Mat{X} \in \mathrm{St}(p,d)$}.
\begin{definition}
A \grassmann manifold $\mathcal{G}(p,d)$ consists of the set of all linear
$p$-dimensional subspaces of $\mathbb{R}^d$.
\end{definition}

The Riemannian inner product, or metric for two tangent vectors $\Delta_1$ and  $\Delta_2$ at $\Mat{X}$
is defined as {\small $\langle \Delta_1, \Delta_2 \rangle_\Mat{X} =
\tr{\Big(\Delta_1^T\big(\mathbf{I}_d - \frac{1}{2}\Mat{X}\Mat{X}^T\big) \Delta_2\Big)} =
\tr{\Big(\Delta_1^T\Delta_2\Big)}$}.%
Further properties of the Riemannian structure of \grassmannian are given in~\cite{Absil:2008}.
This Riemannian structure induces a geodesic distance on the \grassmann, namely the length
of the shortest curve between two points ($p$-dimensional subspaces),
denoted $\delta_g(\Mat{X}_1,\Mat{X}_2)$.
The special orthogonal group \mbox{\small{$SO(d)$}}
(think of this as higher-dimensional rotations)
acts transitively on $\mathcal{G}(p,d)$ by mapping one $p$-dimensional
subspace to another. The geodesic distance may be thought of
as the magnitude of the smallest rotation (element
of {$SO(d)$}) that takes one subspace to the other.
If {$\Theta = [\theta_1, \theta_2, \ldots, \theta_p]$}
is the sequence of principal angles~\cite{Absil:2008} between two subspaces {\small{$\Mat{X}_1$}} and
\mbox{\small{$\Mat{X}_2$}}, then
 \noindent
 \begin{equation}
    \delta_g\left(\Mat{X}_1,\Mat{X}_2\right)=\|\Theta\|_2.
    \label{eqn:geodesic_Grass}
\end{equation}%
\begin{definition}
Let $\Mat{X}_1$ and $\Mat{X}_2$ be two orthonormal matrices of size $d \times p$. The principal angles
$0 \leq \theta_1 $ $\leq \theta_2 \leq $ $\cdots $ $ \leq \theta_p \leq \pi/2$ between two subspaces
$\operatorname{Span}(\Mat{X}_1)$ and $\operatorname{Span}(\Mat{X}_2)$, are defined recursively by
\end{definition}
\begin{eqnarray}
  &\cos(\theta_i)
  =
  \underset{\Vec{u}_i \in \operatorname{Span}(\Mat{X}_1)}{\max}\;
  \underset{\Vec{v}_i \in \operatorname{Span}(\Mat{X}_2)}{\max}\;
  \Vec{u}_i^T \Vec{v}_i  \\
  \text{s.t.:}
  &\|\Vec{u}_i\|_2 \mbox{~=~} \|\Vec{v}_i\|_2 \mbox{~=~} 1  \nonumber\\
  &\Vec{u}_i^T \Vec{u}_j \mbox{~=~} 0;\; j=1,2,\cdots,i-1     \nonumber\\
  &\Vec{v}_i^T \Vec{v}_j \mbox{~=~} 0;\; j=1,2,\cdots,i-1     \nonumber
  \label{eqn:Principal_Angle}
\end{eqnarray}%
In other words, the first principal angle $\theta_1$ is the smallest
angle between all pairs of unit vectors in the first and the second subspaces. The rest of the principal
angles are defined similarly.
The cosines of principal
angles are the singular values of \mbox{$\Mat{X}_1^T \Mat{X}_2$}
\cite{Absil:2008}.

\subsection{Problem statement}
Given a finite set of observations
$\mathbb{X} = \left \{ \Vec{x}_i  \right \}_{i=1}^{m}, \: \Vec{x} \in \mathbb{R}^d$,
dictionary learning in vector spaces optimises the objective function
\begin{equation}
	f(\mathbb{D}) \triangleq \sum\nolimits_{i=1}^{m} l_E(\Vec{x}_i,\mathbb{D})
    \label{eqn:euc_dic_leanring}
\end{equation}
with \mbox{\small$\mathbb{D}_{d \times N} = \left[ \Vec{d}_1 | \Vec{d}_2| \cdots | \Vec{d}_N \right]$}
being a dictionary of size $N$ with atoms \mbox{\small$\Vec{d}_i \in \mathbb{R}^d$}.
Here, $l_E(\Vec{x},\mathbb{D})$ is a loss function and should be small if $\mathbb{D}$ is ``good''
at representing the signal $\Vec{x}$.
Aiming for sparsity, the $ \ell_1 $-norm regularisation is usually
employed to obtain the most common form of $l_E(\Vec{x},\mathbb{D})$ in the literature\footnote{The notation
$[ \cdot ]_i$ and $[\cdot]_{i,j}$ is used to demonstrate elements in position $i$ and $(i,j)$ in a vector and matrix, respectively.}:
\noindent
\begin{equation}
	l_E(\Vec{x},\mathbb{D}) \triangleq
    \underset{\Vec{y}}{\min} \:
    \Bigl\| \Vec{x}- \sum\nolimits_{j=1}^{N} [\Vec{y}]_{j} \Vec{d}_j
    \Bigr\|_2^2
    +\lambda \|\Vec{y}\|_1.
    \label{eqn:euc_sparse_coding}
\end{equation}

With this choice, finding the optimum $\mathbb{D}$ in Eqn.~\eqref{eqn:euc_dic_leanring} is not trivial because of non-convexity,
as can be easily seen by rewriting the dictionary learning to:
\begin{equation*}
\underset{ \{\Vec{y}_i \}_{i=1}^m, \mathbb{D}}{\min} \:
    \sum\nolimits_{i=1}^{m}\left\| \Vec{x}_i- \sum\nolimits_{j=1}^{N} [\Vec{y}_i]_j \Vec{d}_j \right\|_2^2
    +\lambda
    \sum\nolimits_{i=1}^{m} \|\Vec{y}_i\|_1.
\end{equation*}
A common approach to solving this is to alternate
between the two sets of variables, $\mathbb{D}$ and $\Mat{Y} = \left[ \Vec{y}_1 | \Vec{y}_2| \cdots | \Vec{y}_m \right]$, as proposed for example by Aharon \etal~\cite{ELAD_SR_BOOK_2010}.
More specifically, minimising Eqn.~\eqref{eqn:euc_sparse_coding} over sparse codes $\Vec{y}$
while dictionary $\mathbb{D}$ is fixed is a convex problem. Similarly, minimising the overall problem
over $\mathbb{D}$ with fixed $\{\Vec{y}_i \}_{i=1}^m$ is convex as well.

Directly translating the dictionary learning problem
to non-flat \grassmann manifolds results in writing Eqn.~\eqref{eqn:euc_sparse_coding} as:
\begin{equation}
	l_\mathcal{G}(\Mat{X},\mathbb{D}) \triangleq
	\underset{\Vec{y}}{\min} \:
    \Bigl\| \Mat{X} \ominus
    \biguplus\nolimits_{j=1}^{N} [\Vec{y}]_{j} \odot \Mat{D}_j \Bigr\|_\mathcal{G}^2
    +\lambda \|\Vec{y}\|_1.
    \label{eqn:GDL_Grass_sc}
\end{equation}
Here {$\Mat{X} \in \mathbb{R}^ {   d \times p   } $} and {$\Mat{D}_j \in \mathbb{R}^ {   d \times p   }  $}
are points on the \grassmann manifold $\mathcal{G}_{p,d}$,
while the operators $\ominus$ and $\biguplus$  are \grassmann replacements
for subtraction and summation in vector spaces (and hence should be commutative and associative).
Furthermore, $\odot$ is the replacement for scalar multiplication and {$\left\| \cdot \right\|_\mathcal{G}$} is
the {geodesic distance} on \grassmann manifolds.

There are several difficulties in solving Eqn.\ \eqref{eqn:GDL_Grass_sc}.
Firstly, $\ominus$, $\biguplus$ and $\odot$ need to be appropriately defined.
While the Euclidean space is closed under the subtraction and addition
(and hence a sparse solution {\small $\sum_{j=1}^{N}[\Vec{y}]_j\Vec{d}_j$} is a point in that space),
\grassmann manifolds are not closed under normal matrix subtraction and addition.
More importantly, fixing sparse codes $\Vec{y}_i$ does not result in a convex cost function for dictionary learning,
\ie $\sum\nolimits_{i=1}^{m} l_\mathcal{G}(\Mat{X}_i,\mathbb{D})$ is not convex because of
the distance function {$\left\| \cdot \right\|_\mathcal{G}$} in Eqn.~\eqref{eqn:GDL_Grass_sc}.

\section{\grassmann Dictionary Learning (GDL)}
\label{sec:proposed_method}

In this work, we propose to embed \grassmann manifolds into the space of symmetric matrices
via mapping \mbox{\small{$\Pi: \mathcal{G}(p,d) \rightarrow \rm{Sym}(d), \Pi(\Mat{X}) = \Mat{X}\Mat{X}^T$}}.
The embedding {\small $\Pi(\Mat{X})$} is diffeomorphism~\cite{Helmke-Grassmann}
(a one-to-one, continuous, differentiable mapping with a continuous, differentiable inverse)
and has been used for example in subspace tracking~\cite{Srivastava_2004}.
It has been used in~\cite{Cetingul:CVPR:2009} for clustering
and in~\cite{HAMM2008_ICML} and \cite{Harandi_CVPR_2011} to develop discriminant analysis
on \grassmann manifolds amongst the others.
The induced space can be understood as a smooth, compact submanifold of {\small{$\rm{Sym}(d)$}}
of dimension $d(d - p)$~\cite{Helmke-Grassmann}.
A natural metric on $\rm{Sym}(d)$ is induced by the Frobenius norm, $\|\Mat{A}\|_F^2 = \tr(\Mat{A}\Mat{A}^T)$ which
we will exploit to {\em convexify}  Eqn.~\eqref{eqn:GDL_Grass_sc}.
Here $ \tr(\cdot) $ is the matrix trace operator.
As such, we define:
\[
    \delta_s(\Mat{X}_1,\Mat{X}_2) = \|\Pi(\Mat{X}_1) - \Pi(\Mat{X}_2)\|_F = \|\Mat{X}_1\Mat{X}_1^T - \Mat{X}_2\Mat{X}_2^T\|_F
\]
as the metric in the induced space.
Before explaining our solution,
we note that $\delta_s$ is related to the \grassmann manifold
in several aspects. This provides motivation and grounding for the
follow-up formulation and is discussed briefly here.
\begin{theorem} \label{thm:isometry_thm}
The mapping $\Pi(\Mat{X})$ forms an isometry from
{ $\left(\mathcal{G}(p,d),\delta_p \right)$}
onto the {$\left( \rm{Sym}(d), \delta_s\right)$}
where the {\it projection distance} between two points on the \grassmann manifold {$\mathcal{G}(p,d)$}
is defined as
{ $\delta_p^2\left(\Mat{X}_1,\Mat{X}_2\right) =  \sum_{i=1}^p\sin^2\theta_i$}.
\end{theorem}
We refer the reader to~\cite{HAMM2008_ICML} for the proof of this theorem.
\begin{theorem} \label{thm:curve_equiv_thm}
\label{thm:intrinsic_metric}
The length of any given curve is the same under $\delta_s$ and
$\delta_g$ up to a scale of $\sqrt{2}$.
\end{theorem}
The proof of this theorem is in the Appendix.

Since {\small $\Pi(\Mat{X})$} is in the manifold of symmetric matrices,
matrix subtraction and addition can be considered for $\ominus$ and $\biguplus$ in Eqn.~\eqref{eqn:GDL_Grass_sc}.
Therefore, we recast the dictionary learning task as optimising the empirical cost function
$f(\mathbb{D}) \triangleq \sum\nolimits_{i=1}^{m} l(\Mat{X}_i,\mathbb{D})$
with
\begin{equation}
    l(\Mat{X},\mathbb{D}) \triangleq
    \underset{\Vec{y}}{\min} \:
    \Bigl\| \Mat{X}\Mat{X}^T -
    \sum_{j=1}^{N} [\Vec{y}]_{j} \Mat{D}_j\Mat{D}_j^T \Bigr\|_F^2
    +\lambda \|\Vec{y}\|_1.
    \label{eqn:GDL_Grass}
\end{equation}%

It is this projection mapping
$\Pi( \Mat{X} ) = \Mat{X} \Mat{X}^ T$ that leads to a simple and efficient learning approach to our
problem.

\subsection{Sparse Coding}
Finding the sparse codes with fixed $\mathbb{D}$ is straightforward, as
expanding the Frobenius norm term in Eqn.~\eqref{eqn:GDL_Grass} results in a convex function in $\Vec{y}$:
\begin{align*}
  &
    \Bigl\| \Mat{X}\Mat{X}^T -    \sum_{j=1}^{N} [\Vec{y}]_j
    \Mat{D}_j\Mat{D}_j^T \Bigr\|_F^2
    =
    \tr(\Mat{X}^T\Mat{X}\Mat{X}^T\Mat{X})   +
    \nonumber\\
  &  \!  \sum_{j,r =1}^{N}{[\Vec{y}]_j [\Vec{y}]_r
\tr(\Mat{D}_r^T\Mat{D}_j\Mat{D}_j^T\Mat{D}_r)}                          %
	-2\sum_{j=1}^{N}{[\Vec{y}]_j ~ \tr(\Mat{D}_j^T\Mat{X}\Mat{X}^T\Mat{D}_j)}.
\end{align*}%

\noindent
Sparse codes can be obtained without explicit embedding of the manifold to
$\rm{Sym}(d)$ using $\Pi(\Mat{X})$. This can be seen by defining
\mbox{$[\mathcal{K}(\Mat{X},\mathbb{D})]_i = \tr(\Mat{D}_i^T\Mat{X}\Mat{X}^T\Mat{D}_i) = \|\Mat{X}^T\Mat{D}_i\|_F^2$}
as an $N$ dimensional vector storing  the similarity between signal $\Mat{X}$
and dictionary atoms in the induced space and
{\small $[\mathbb{K}(\mathbb{D})]_{i,j} =
 \tr(\Mat{D}_i^T\Mat{D}_j\Mat{D}_j^T\Mat{D}_i) = \|\Mat{D}_i^T\Mat{D}_j\|_F^2$}
as an {\small $N \times N$} symmetric matrix encoding the similarities between dictionary atoms (which can be computed offline).
Then, the sparse coding in  Eqn.~\eqref{eqn:GDL_Grass} can be written as:
\begin{equation}
	l(\Mat{X},\mathbb{D}) = \underset{\Vec{y}}{\min} \:
	\Vec{y}^T \mathbb{K}(\mathbb{D}) \Vec{y} -2\Vec{y}^T\mathcal{K}(\Mat{X},\mathbb{D})
	+\lambda \|\Vec{y}\|_1,
	\label{eqn:Opt_Grass3}
\end{equation}
which is a quadratic problem. Clearly
the symmetric matrix $  \mathbb{K}(\mathbb{D})  $ is positive
semidefinite. So the  quadratic  problem is convex.

\subsection{Dictionary Update}
To update dictionary atoms we break the minimisation problem into $N$ sub-minimisation problems by independently updating each atom,
in line with general practice in dictionary learning~\cite{ELAD_SR_BOOK_2010}.
More specifically, fixing sparse codes and ignoring the terms that are irrelevant to dictionary atoms,
the dictionary learning problem can be seen as finding
{{$ \underset{\mathbb{D}}{\min} \sum_{r=1}^{N}{\mathcal{J}(r)}$}}, where:
\begin{eqnarray}
    \mathcal{J}(r) &= &\sum\nolimits_{i=1}^{m}{
    \sum\nolimits_{j=1, j \neq r}^{N}{[\Vec{y}_i]_{r}[\Vec{y}_i]_{j} \tr( \Mat{D}_r^T\Mat{D}_j\Mat{D}_j^T\Mat{D}_r )}} \nonumber\\
    & &-2\sum\nolimits_{i=1}^{m}{[\Vec{y}_i]_{r} \tr(
    \Mat{D}_r^T\Mat{X}_i\Mat{X}_i^T\Mat{D}_r)}.
    \label{eqn:dic_sub_probl}
\end{eqnarray}%
Imposing the orthogonality of $\Mat{D}_r$ results in the following minimisation sub-problem for updating $\Mat{D}_r$:
\begin{equation}
    \Mat{D}_r^\ast = \underset{\Mat{D}_r}{\operatorname{argmin}} \;
    \mathcal{J}(r),  ~~~ \text{s.t.} ~~~ \Mat{D}_r^T\Mat{D}_r =
    \mathbf{I}_p.
    \label{eqn:GDL_opt1}
\end{equation}%
A closed-form solution for the above minimisation problem can be obtained by the method of Lagrange multipliers
and forming
\mbox{\small $L(r,\zeta) = \mathcal{J}(r) +
\zeta\left(\Mat{D}_r^T\Mat{D}_r - \mathbf{I}_p\right)$}.
The gradient of {\small $L(r,\zeta)$} is:
\begin{align}
    \nabla_{\Mat{D}_r}L(r,\zeta) &=  2\sum\nolimits_{i=1}^{m}{\sum\nolimits_{j=1, j \neq r}^{N}{[\Vec{y}]_{r}[\Vec{y}]_{j}\Mat{D}_j\Mat{D}_j^T\Mat{D}_r}} \nonumber\\
    &  -4\sum\nolimits_{i=1}^{m}{[\Vec{y}_i]_{r}\Mat{X}_i\Mat{X}_i^T\Mat{D}_r}
    +2\zeta \Mat{D}_r.
    \label{eqn:proj_lagrange2}
\end{align}%
The solution of Eqn.\ \eqref{eqn:GDL_opt1} can be sought by finding the roots of \eqref{eqn:proj_lagrange2},
\ie, {\small $\nabla_{\Mat{D}_r}L(r,\zeta) = 0$},
which is an eigen-value problem.
Therefore, the solution of \eqref{eqn:GDL_opt1} can be obtained by computing $p$ eigenvectors of $\mathcal{S}$, where
\begin{equation}
    \mathcal{S} = \sum_{i=1}^{m}{ \sum_{j=1, j \neq r}^{N}{[\Vec{y}_i]_{r}[\Vec{y}_i]_{j}\Mat{D}_j\Mat{D}_j^T}  }
    - 2\sum_{i=1}^{m}{[\Vec{y}_i]_{r}\Mat{X}_i\Mat{X}_i^T}.
    \label{eqn:proj_lagrange_sol}
\end{equation}%

Note that the mapping $\Pi(\Mat{X})$ meets the requirement of a \grassmann kernel~\cite{HAMM2008_ICML,Harandi_PRL_2013}.
Consequently, it is possible to interpret the above solution as a kernel method.
Nevertheless, we believe that the explanation through manifold of symmetric matrices provides more insight into the solution
and also avoids possible confusion with the following section,
where we present an explicitly kernelised version of GDL.

\section{Kernelised GDL}
\label{sec:kernel_gdl}

In this section we propose the kernel extension of the GDL method (KGDL) to
model complex nonlinear structures in the original data.

Assume a population of sets in the form of
{ $\mathbb{X} = $ $ \left\{  \Mat{X}_i \right\}_{i=1}^{m}$},
with $\Mat{X}_i = $ $ \left\{  \Vec{x}_{j}^{i} \right\}_{j=1}^{m_i}$;
$ \Vec{x}_{j}^{i} \in \mathbb{R}^d$
and a kernel function $ k( \cdot, \cdot ) $ is given.
Therefore, a real-valued function on {$\mathbb{R}^d \times \mathbb{R}^d$}
with the property that a mapping {$\phi : \mathbb{R}^d \rightarrow \mathcal{H}$}
into a dot product Hilbert space $\mathcal{H}$ exists,
such that for all {$\Vec{x},\Vec{x}^\prime \in \mathbb{R}^d$}
we have
{$\langle\Vec{x},\Vec{x}^\prime\rangle_\mathcal{H} = \phi(\Vec{x})^T \phi(\Vec{x}^\prime) = k(\Vec{x},\Vec{x}^\prime)$}
\cite{Shawe-Taylor:2004:KMP}.
As before, we are interested in optimising
{\small$f(\mathbb{D}) \triangleq \sum\nolimits_{i=1}^{m} l_\Psi(\Mat{X}_i,\mathbb{D})$}
where $l_\Psi(\Mat{X},\mathbb{D})$ is the kernel version of
\eqref{eqn:GDL_Grass} as depicted below:

\noindent
\begin{align}
	& l_\Psi(\Mat{X},\mathbb{D}) \triangleq
    \notag \\
    &
    \underset{\Vec{y}}{\min} \:
    \Bigl \| \Psi(\Mat{X})\Psi(\Mat{X})^T \hspace{-1ex} - \hspace{-1ex}
    \sum_{j=1}^{N} [\Vec{y}]_{j} \Psi(\Mat{D}_j)\Psi(\Mat{D}_j)^T
    \Bigr\|_F^2
    +\lambda \|\Vec{y}\|_1.
    \label{eqn:KGDL_Grass}
\end{align}%

Here, {$\Psi(\Mat{Z}) = [\Vec{\psi}_{1}|\Vec{\psi}_{2}|\cdots|\Vec{\psi}_{p}]$}
denotes a subspace of order $p$ in the space $\mathcal{H}$
associated to samples of set \mbox{$\Mat{Z} = \{\Vec{z}_{i} \}_{i=1}^{m_z}$}.
That is,
{$\Vec{\psi}_{j} = \sum_{i=1}^{m_z} a_{i,j} \phi(\Vec{z}_{i})$}
and \mbox{$\Psi(\Mat{Z}) = \Mat{\Phi}(\Mat{Z})\Mat{A}_{\Mat{Z}}$}
with {$[\Mat{A}_{\bf Z}]_{i,j} = a_{i,j}$}.

We note that the coefficients $a_{i,j}$ are given by the KPCA~\cite{Shawe-Taylor:2004:KMP} method for input sets (\ie, $\Mat{X}_i$)
while in the case of the dictionary atoms $\Mat{D}_j$,  they need to be determined by the KGDL algorithm.
We also note that
{$\Psi(\Mat{Z})^T\Psi(\Mat{W}) = \Mat{A}_Z^T \Mat{K}(\Mat{Z},\Mat{W})\Mat{A}_W$}
with \mbox{$\Mat{K}(\Mat{Z},\Mat{W})$} being the kernel matrix of size $m_z \times m_w$
between sets $\Mat{Z}$ and $\Mat{W}$, \ie,
{$[\Mat{K}(\Mat{Z},\Mat{W})]_{i,j} = \phi(\Vec{z}_{i})^T\phi(\Vec{w}_{j}) =k(\Vec{z}_{i},\Vec{w}_{j})$}.

Obtaining sparse codes $\Vec{y}$ is a straightforward task since
\mbox{\small $\left\| \Psi(\Mat{X})\Psi(\Mat{X})^T - \sum_{j=1}^{N} [\Vec{y}]_j \Psi(\Mat{D}_j)\Psi(\Mat{D}_j)^T \right\|_F^2$}
is a convex function in $\Vec{y}$.
Similar to the linear GDL method, dictionary is updated atom by atom (\ie, atoms are assumed to be independent) by
fixing sparse codes.
As such, the cost function to update {$\Psi(\Mat{D}_r)$} can be defined as:
\begin{eqnarray}
    \mathcal{J}_\Psi(r) \hspace{-2ex}&=&\hspace{-2ex} \sum_{i,j}
    {[\Vec{y}_i]_{r}[\Vec{y}_i]_{j} \tr( \Psi(\Mat{D}_r)^T\Psi(\Mat{D}_j)\Psi(\Mat{D}_j)^T
    \Psi(\Mat{D}_r) )} \nonumber\\
    \hspace{-2ex}& &\hspace{-2ex}\hspace{-3ex}-2\sum_{i=1}^{m}{[\Vec{y}_i]_{r}
    \tr( \Psi(\Mat{D}_r)^T\Psi(\Mat{X}_i)\Psi(\Mat{X}_i)^T\Psi(\Mat{D}_r) )},\nonumber\\
    \hspace{-2ex}&=&\hspace{-2ex} \sum_{i=1}^{m}{\sum_{j=1, j \neq r}^{N}{[\Vec{y}_i]_{r}[\Vec{y}_i]_{j}
    \tr(\Mat{A}_{\Mat{D}_r}^T\Mat{B}(\Mat{D}_r,\Mat{D}_j)\Mat{A}_{\Mat{D}_r})}}\nonumber\\
    \hspace{-2ex}& &\hspace{-3ex}-2\sum_{i=1}^{m}{[\Vec{y}_i]_{r}
    \tr(    \Mat{A}_{\Mat{D}_r}^T\Mat{B}(\Mat{D}_r,\Mat{X}_i)\Mat{A}_{\Mat{D}_r}
    )}.
    \label{eqn:kdic_sub_probl}
\end{eqnarray}%
where
{$\Mat{B}(\Mat{X},\Mat{Z}) = \Mat{K}(\Mat{X},\Mat{Z})\Mat{A}_{\bf Z}\Mat{A}_{\bf Z}^T \Mat{K}(\Mat{Z},\Mat{X})$}.
The orthogonality constraint in $\mathcal{H}$ can be written as:
\begin{equation}
    \Psi(\Mat{D}_r)^T\Psi(\Mat{D}_r) =
    \Mat{A}_{\Mat{D}_r}^T\Mat{K}(\Mat{D}_r,\Mat{D}_r)\Mat{A}_{\Mat{D}_r}
    = \mathbf{I}_p.
    \label{eqn:KGDL_orth_condition}
\end{equation}%
{\small $\Psi(\Mat{D}_r)$} is fully determined if
{\small{$\Mat{A}_{\Mat{D}_r}$}} and \mbox{\small{$\Mat{K}(\cdot,\Mat{D}_r)$}} are known.
If we assume that dictionary atoms are independent,
then according to the representer theorem~\cite{Shawe-Taylor:2004:KMP},
\mbox{\small $\Psi(\Mat{D}_r)$} is a linear combination of all the {\small{$\Mat{X}_i$}} that have used it.
Therefore, \mbox{\small $\Mat{K}(\cdot,\Mat{D}_r) = \Mat{K}(\cdot,\bigcup_{i}\Mat{X}_i), \; y_{i,r} \neq 0$}
which leaves us with identifying {$\Mat{A}_{\Mat{D}_r}$}
via the following minimisation problem:
\begin{eqnarray*}
    &\underset{\Mat{A}_{D_r}}{\min}
    &\sum\limits_{i=1}^{m}{\sum\limits_{j=1, j \neq r}^{N}{y_{i,r}y_{i,j}
    \tr( \Mat{A}_{\Mat{D}_r}^T\Mat{B}(\Mat{D}_r,\Mat{D}_j)\Mat{A}_{\Mat{D}_r})}}\nonumber\\
    & &-2\sum\nolimits_{i=1}^{m}{y_{i,r}
    \tr( \Mat{A}_{\Mat{D}_r}^T\Mat{B}(\Mat{D}_r,\Mat{X}_i)\Mat{A}_{\Mat{D}_r} )} \nonumber\\
    &\text{s.t.}
    &\Mat{A}_{\Mat{D}_r}^T\Mat{K}(\Mat{D}_r,\Mat{D}_r)\Mat{A}_{\Mat{D}_r}
    = \mathbf{I}_p.
\end{eqnarray*}%
The solution of the above problem is given by the
eigenvectors of the generalised eigenvalue problem
\mbox{$\mathcal{S}_\Psi \Vec{v} = \lambda \Mat{K}(\Mat{D}_r,\Mat{D}_r) \Vec{v}$},
where:
\begin{equation}
    \mathcal{S}_\Psi = \sum_{i=1}^{m}{[\Vec{y}_i]_{r} \Bigl( \sum_{j=1, j \neq r}^{N}{[\Vec{y}_i]_{j}\Mat{B}(\Mat{D}_r,\Mat{D}_j)}
    - 2\Mat{B}(\Mat{D}_r,\Mat{X}_i)\Bigr)}.
    \label{eqn:kgdl_sol}
\end{equation}%

\section{Further Discussion}
The solution proposed in~\eqref{eqn:GDL_Grass} considers $\Pi( \Mat{X} )$ as a mapping and solves sparse coding extrinsically, meaning
$\sum_i [\Vec{y}]_{i} \Mat{D}_i\Mat{D}_i^T$ is not necessarily a point on $\mathcal{G}(p,d)$.
If, however, it is required that the linear combination of elements
$\sum_i [\Vec{y}]_i \Mat{D}_i \Mat{D}_i^T$ actually be used to represent a point on the
Grassmannian, it can be accomplished as follows.
The \textbf{Eckart-Young} theorem~\cite{Golub_Book} states that
the matrix of rank $p$ closest in Frobenius norm to a given matrix $\Mat{X}$
is found by dropping all the singular values beyond the $p$-th one.
This operation (along with equalization of the singular values)
can easily be applied to a linear combination of
matrices $\Mat{D}_i \Mat{D}_i^T$ to obtain a point on the Grassmann manifold.
Thus, in a very concrete sense, the linear combination of elements
$\Mat{D}_i \Mat{D}_i^T$, although not equaling any point on the Grassmann manifold, does
{\em represent} such an element, the closest point lying on the manifold
itself.

Furthermore,~\eqref{eqn:GDL_Grass} follows the general principle of sparse coding in that the over-completeness of $\mathbb{D}$ will  approximate $\Mat{X}\Mat{X}^T$ and $\sum_i [\Vec{y}]_i \Mat{D}_i \Mat{D}_i^T$ could be safely expected to be closely tied to a Grassmannian point.
Since $d \times d$ symmetric matrices of rank $p$ with the extra property of being idempotent\footnote{%
A matrix $\Mat{P}$ is called idempotent if $\Mat{P}^2 = \Mat{P}$.}
are equivalent to points on $\mathcal{G}(p,d)$
, an intrinsic version of~\eqref{eqn:GDL_Grass} can be written as:
\begin{equation}
    \underset{\Vec{y}}{\min} \:
    \Bigl\| \Mat{X}\Mat{X}^T -
    \mathrm{Proj}\Big(\sum_{j=1}^{N} [\Vec{y}]_{j} \Mat{D}_j\Mat{D}_j^T\Big) \Bigr\|_F^2
    +\lambda \|\Vec{y}\|_1,
    \label{eqn:rene_equ}
\end{equation}%
where $\mathrm{Proj}(\cdot)$ is the operator that projects  a symmetric matrix onto a Grassmann manifold (by forcing the idempotency and rank properties). Based on Eckart-Young theorem,
$\mathrm{Proj}(\cdot)$ can be obtained through Singular Value Decomposition (SVD).
The involvement of SVD, especially in vision applications, makes solving~\eqref{eqn:rene_equ} tedious and challenging.
We acknowledge that seeking efficient ways of solving~\eqref{eqn:rene_equ} is interesting but beyond this paper.

\section{Experiments}
\label{sec:experiments}

In this section we compare and contrast the performance of the proposed GDL and KGDL methods
against several state-of-the-art methods:
Discriminant analysis of Canonical Correlation (DCC)~\cite{DCCA:PAMI:2007},
kernel version of Affine Hull Method (KAHM)~\cite{Cevikalp_CVPR_2010},
Grassmann Discriminant Analysis (GDA)~\cite{HAMM2008_ICML},
and Graph-embedding Grassmann Discriminant Analysis (GGDA)~\cite{Harandi_CVPR_2011}.
We evaluate the performance on the tasks of
\textbf{(i)} face recognition from image sets,
\textbf{(ii)} dynamic texture classification and
\textbf{(iii)} action recognition.

DCC is an iterative learning method that maximises a measure of discrimination between image sets
where the distance between sets is expressed by canonical correlations.
In KAHM, images are considered as points in a linear or affine feature space,
while image sets are characterised by a convex geometric region (affine or convex hull) spanned by their feature points.
GDA can be considered as an extension of kernel discriminant analysis over Grassmann manifolds \cite{HAMM2008_ICML}.
In GDA, a transform over the Grassmann manifold is learned
to simultaneously maximise a measure of inter-class distances and minimise intra-class distances.
GGDA can be considered as an extension of GDA,
where a local discriminant transform over Grassmann manifolds is learned.
This is achieved by incorporating local similarities/dissimilarities through within-class and between-class similarity graphs.

Based on preliminary experiments, the Gaussian kernel~\cite{Shawe-Taylor:2004:KMP}
was used in KGDL for all tests.
In GDL and KGDL methods, the dictionary is used to generate sparse codes for training and testing data.
The sparse codes are then fed to a SVM for classification.
The size of the dictionary is found empirically and the highest accuracy is reported here.
In all experiments, the input data has the form of image sets. An image set
\mbox{$\mathbb{F} = \left\{ \Vec{f}_i \right\}_{i=1}^{b};\Vec{f}_i \in \mathbb{R}^d$},
with \mbox{$\Vec{f}_i$} being the vectorised descriptor of frame $i$,
can be modelled by a subspace (and hence as a point on a \grassmann manifold)
through any orthogonalisation procedure like SVD.
More specifically,
let \mbox{$\mathbb{F} = \Mat{U} \Mat{D} \Mat{V}^T$} be the SVD of $\mathbb{F}$.
The dominant~$p$ left singular-vectors ($p$ columns of~$\Mat{U}$ corresponding to
the maximum singular values) represent
an optimised subspace of order~$p$ (in the mean square sense) for~$\mathbb{F}$
and can be seen as a point on manifold~{\small $\mathcal{G}{p,d}$}.
To select the optimum value of $p$,  (order) of subspaces, the performance of a NN classifier on Grassmann manifolds will be evaluated for various values of `p' and the value resulted in maximum performance will be chosen for constructing subspaces for each task.

\subsection{Face Recognition}
\label{sec:face_recognition}

While face recognition from a single still image has been extensively studied,
recognition based on a group of still images is relatively new.
A popular choice for modelling image-sets
is by representing them through linear subspaces \cite{HAMM2008_ICML,Harandi_CVPR_2011}.
For the task of image-set face recognition, we used the YouTube celebrity dataset \cite{YT_Celebrity}.
See Fig.~\ref{fig:YT_Celebrity_example} for examples.
Face recognition on this dataset is very challenging,
since the videos are compressed with a very high compression ratio and most of them are low-resolution.

To create an image set from a video, we used a cascaded face detector \cite{Viola:IJCV:2004}
to extract face regions from each video, followed by
resizing regions to $96 \times 96$ and describing them via histogram of Local Binary Patterns
(LBP)~\cite{LBP_PAMI_2002}.
Then each image set (corresponding to a video) was represented by a linear subspace of order $5$.
Our data included 1471 image-sets which ware randomly split into 1236 training and 235 testing points.
The process of random splitting was repeated ten times and the average classification accuracy is reported.
The results in Table \ref{tab:table_face_rec} show that the proposed
GDL and KGDL approaches outperform the competitors. KGDL achieved the highest accuracy of 73.91, more than 3
percentage points better than GDL.

\begin{table}[!b]
  \centering
  \caption
    {
    Average recognition rate on the YouTube celebrity dataset.
    }
  \label{tab:table_face_rec}
  \begin{tabular}{lccccc}
    \toprule
    \bf{Method}    &\bf{CRR}  \\
    \toprule
    {DCC~\cite{DCCA:PAMI:2007}}      			  &$60.21 \pm 2.9$ \\
    {KAHM~\cite{Cevikalp_CVPR_2010}}              &$67.49 \pm 3.5$ \\
    {GDA~\cite{HAMM2008_ICML}}                    &$58.72 \pm 3.0$\\
    {GGDA~\cite{Harandi_CVPR_2011}}               &$61.06 \pm 2.2$\\
    {\bf{GDL}}                                    &$70.47 \pm 1.7$\\
    {\bf{KGDL}}                                   &$\bf{73.91 \pm 1.9}$\\
    \bottomrule
  \end{tabular}
\end{table}

\def \YTSIZE {0.2}
\begin{figure}[!tb]
  \begin{minipage}{1\columnwidth}\center
  \includegraphics[width=\YTSIZE \columnwidth,keepaspectratio]{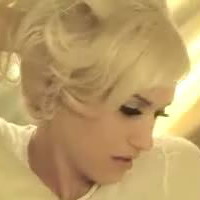}
  \includegraphics[width=\YTSIZE \columnwidth,keepaspectratio]{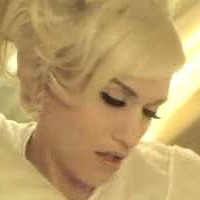}
  \includegraphics[width=\YTSIZE \columnwidth,keepaspectratio]{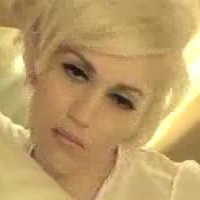}
  \includegraphics[width=\YTSIZE \columnwidth,keepaspectratio]{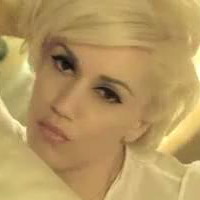}
  \end{minipage}
  \begin{minipage}{1\columnwidth}\center
  \includegraphics[width=\YTSIZE \columnwidth,keepaspectratio]{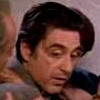}
  \includegraphics[width=\YTSIZE \columnwidth,keepaspectratio]{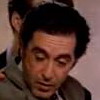}
  \includegraphics[width=\YTSIZE \columnwidth,keepaspectratio]{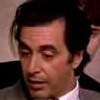}
  \includegraphics[width=\YTSIZE \columnwidth,keepaspectratio]{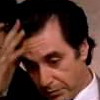}
  \end{minipage}
  \caption
    {
    Examples of YouTube celebrity dataset (grayscale versions of images were used in our experiments).
    }
  \label{fig:YT_Celebrity_example}
\end{figure}

\subsection{Dynamic Texture Classification}
\label{sec:dynamic_texture}

Dynamic textures are videos of moving scenes
that exhibit certain stationary properties in the time domain~\cite{DynTex_Dataset}.%
Such videos are pervasive in various environments,
such as sequences of rivers, clouds, fire, swarms of birds, humans in crowds.
In our experiment, we used the challenging DynTex++ dataset~\cite{DynTex_Dataset},
which is comprised of 36 classes,
each of which contains 100 sequences with a fixed size of $50 \times 50 \times 50$
(see Fig.~\ref{fig:DynTex_example} for example classes).
We split the dataset into training and testing sets by randomly assigning half of the videos of each class to
the training set and using the rest as query data.
The random split was repeated twenty times; average accuracy is reported.

To generate points on the \grassmann manifold,
we used histogram of LBP from Three Orthogonal Planes (LBP-TOP)~\cite{LBPTOP:PAMI:2007} which,
takes into account the dynamics within the videos. To this end, we split each video
to subvideos of length 10, with a 7 frames overlap. Each subvideo then described by a histogram of LBP-TOP
features. From the subvideo descriptors, we extracted a subspace of order 5 as
the video representation on \grassmann manifold.

In addition to DCC~\cite{DCCA:PAMI:2007}, KAHM~\cite{Cevikalp_CVPR_2010}, GDA~\cite{HAMM2008_ICML} and GGDA~\cite{Harandi_CVPR_2011},
the proposed GDL and KGDL approaches were compared against
Distance Learning Pegasos (DL-Pegasos)~\cite{DynTex_Dataset}.
DFS can be seen as concatenation of two components:
(i)~a volumetric component that encodes the stochastic self-similarities of dynamic textures as 3D volumes,
(ii)~a multi-slice dynamic component that captures structures of dynamic textures on 2D slices along various views of the 3D volume.
DL-Pegasos uses three descriptors (LBP, HOG and LDS)
and learns how the descriptors can be linearly combined to best discriminate between dynamic texture classes.

The overall classification results are presented in Table~\ref{tab:table_dyntex}.
The proposed KGDL approach obtains the highest average recognition rate. To our best knowledge this is the
highest reported result on DynTex++ dataset.

\def \DTSIZE {0.225}
\begin{figure}[!tb]
  \centering
  \begin{minipage}{1\columnwidth}
  \includegraphics[width=\DTSIZE \columnwidth,keepaspectratio]{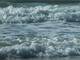}
  \includegraphics[width=\DTSIZE \columnwidth,keepaspectratio]{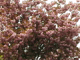}
  \includegraphics[width=\DTSIZE \columnwidth,keepaspectratio]{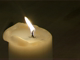}
  \includegraphics[width=\DTSIZE \columnwidth,keepaspectratio]{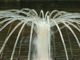}
  \end{minipage}
  \begin{minipage}{1\columnwidth}
  \includegraphics[width=\DTSIZE \columnwidth,keepaspectratio]{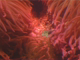}
  \includegraphics[width=\DTSIZE \columnwidth,keepaspectratio]{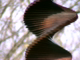}
  \includegraphics[width=\DTSIZE \columnwidth,keepaspectratio]{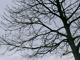}
  \includegraphics[width=\DTSIZE \columnwidth,keepaspectratio]{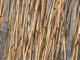}
  \end{minipage}
  \vspace{1ex}
  \caption
    {
    Example classes of DynTex++ dataset (grayscale images were used in our experiment).
    }
  \label{fig:DynTex_example}
\end{figure}

\begin{table}[!tb]
  \centering
  \caption
    {
    Average recognition rate on the DynTex++ dataset.
    }
  \label{tab:table_dyntex}
  \vspace{-1ex}
  \begin{tabular}{lccccc}
    \toprule
    \bf{Method}    &\bf{CRR}  \\
    \toprule
    {DL-PEGASOS~\cite{DynTex_Dataset}}            &$63.7$\\
    {DCC~\cite{DCCA:PAMI:2007}}				  &$53.2$ \\
    {KAHM~\cite{Cevikalp_CVPR_2010}}              &$82.8$ \\
    {GDA~\cite{HAMM2008_ICML}}                    &$81.2$\\
    {GGDA~\cite{Harandi_CVPR_2011}}               &$84.1$\\
    {GDL}	                                      &$90.3$\\
    {\bf{KGDL}}                                   &$\bf{92.8}$\\
    \bottomrule
  \end{tabular}
\end{table}

\subsection{Ballet Dataset}
\label{exp_Ballet}

The Ballet dataset contains 44 videos of 8 actions collected from an instructional ballet DVD~\cite{Ballet_Dataset}%
.
The dataset consists of 8 complex motion patterns performed by 3 subjects,
and is very challenging due to the significant intra-class variations in terms of speed,
spatial and temporal scale, clothing and movement.

We extracted 2400 image sets by grouping every 6 frames that exhibited the same action into one image set.
We descried each image set by a subspace of order 4 with Histogram of Oriented Gradient (HOG) as frame descriptor~\cite{Dalal:2005:HOG}.
Available samples were randomly split into training and testing set (the number of
image sets in both sets was even).
The process of random splitting was repeated
ten times and the average classification accuracy is reported.

Table~\ref{tab:table_ballet} shows that both GDL and KGDL algorithms
have superior performance as compared to the state-of-the-art methods
DCC, KAHM, GDA and KGDA. The difference between KGDL and the closest
state-of-the-art competitor, \ie, GGDA, is roughly ten percentage points.

\begin{table}[!b]
  \centering
  \caption
    {
    Average recognition rate on the Ballet dataset.
    }
  \label{tab:table_ballet}
	\vspace{-1ex}
  \begin{tabular}{lccccc}
    \toprule
    \bf{Method}    &\bf{CRR}  \\
    \toprule
    {DCC~\cite{DCCA:PAMI:2007}}            	  	  &$41.95 \pm 9.6$ \\
    {KAHM~\cite{Cevikalp_CVPR_2010}}              &$70.05 \pm 0.9$ \\
    {GDA~\cite{HAMM2008_ICML}}                    &$67.33 \pm 1.1$\\
    {GGDA~\cite{Harandi_CVPR_2011}}               &$73.54 \pm 2.0$\\
    {\bf{GDL}}                                    &$79.64 \pm 1.1$\\
    {\bf{KGDL}}                                   &$\bf{83.53 \pm 0.8}$\\
    \bottomrule
  \end{tabular}
\end{table}

\def\BALLETSIZE {0.2}
\begin{figure}[!tb]
      \begin{minipage}{1.0\columnwidth}
      \center
        \includegraphics[width=\BALLETSIZE\columnwidth,keepaspectratio]{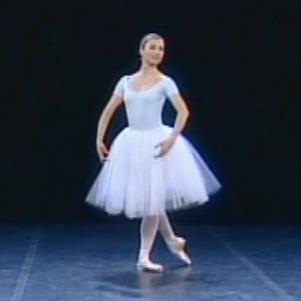}
        \includegraphics[width=\BALLETSIZE\columnwidth,keepaspectratio]{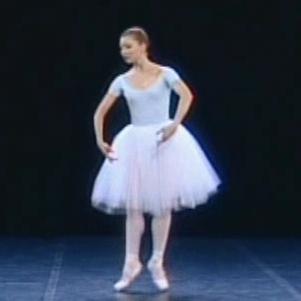}
        \includegraphics[width=\BALLETSIZE\columnwidth,keepaspectratio]{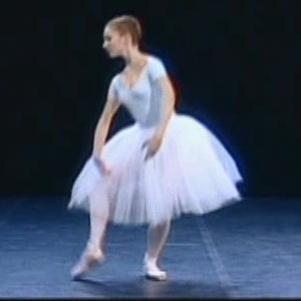}
        \includegraphics[width=\BALLETSIZE\columnwidth,keepaspectratio]{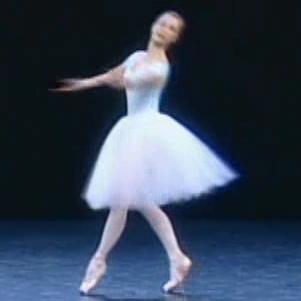}
      \end{minipage}
      \begin{minipage}{1.0\columnwidth}
      \center
        \includegraphics[width=\BALLETSIZE\columnwidth,keepaspectratio]{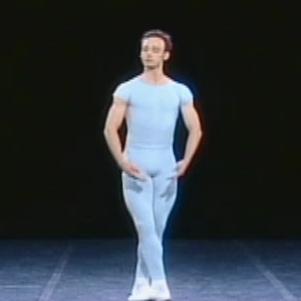}
        \includegraphics[width=\BALLETSIZE\columnwidth,keepaspectratio]{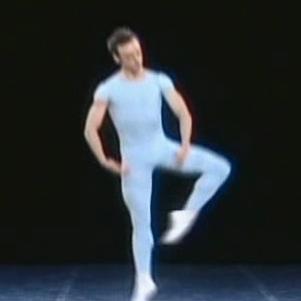}
        \includegraphics[width=\BALLETSIZE\columnwidth,keepaspectratio]{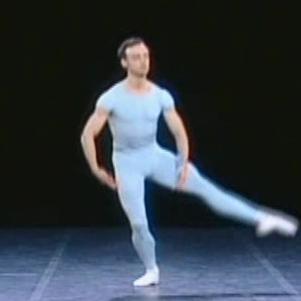}
        \includegraphics[width=\BALLETSIZE\columnwidth,keepaspectratio]{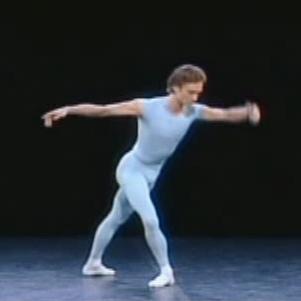}
      \end{minipage}
  \caption
    {
    Examples from the Ballet dataset~\cite{Ballet_Dataset}.
    }
  \label{fig:Ballet_example}
\end{figure}

Please note that in all our experiments, we randomly initialized the dictionary ten times and picked the one with minimum reconstruction error over the training set. We performed an extra experiment on ONE SPLIT of YT dataset and evaluated 10 random initialisations. The mean and std for the GDL were 72.21\% and 1.6 respectively, while the performance for the dictionary with min reconstruction error was observed to be 73.19\%.

Regarding the intrinsic approach which appreciates the geodesic distance in deriving sparse codes, we performed an extra experiment on YT, following~\cite{Vemuri_MICCAI_2012}. To learn the dictionary, the intrinsic approach  required 26706s on an i7-Quad core platform as compared to 955sec for our algorithm. In terms of performance, our algorithm outperformed the intrinsic method (70.47\% as compared to 68.51\%). While this sounds counter-intuitive, we conjecture this might be due to the affine constraint required by the intrinsic approach for generating sparse codes.

\section{Main Findings and Future Directions}
\label{sec:conclusion}
\vspace{0.5ex}

With the aim of learning a Grassmann dictionary,
we first proposed to embed Grassmann manifolds into the space of symmetric matrices by an isometric projection.
We have then shown how sparse codes can be determined in the induced space
and devised a closed-form solution for updating a Grassmann dictionary, atom by atom.
Finally, we proposed a kernelised version of the dictionary learning algorithm,
to handle non-linearity in data.

Experiments on several classification tasks
(face recognition from image sets, action recognition and dynamic texture classification)
show that the proposed approaches achieve notable improvements in discrimination accuracy,
in comparison to state-of-the-art methods such as affine hull method~\cite{Cevikalp_CVPR_2010},
Grassmann discriminant analysis (GDA)~\cite{HAMM2008_ICML},
and graph-embedding Grassmann discriminant analysis~\cite{Harandi_CVPR_2011}.

In this work a Grassmann dictionary is learned such that a reconstruction error is minimised.
This is not necessarily the optimum solution when labelled data is available.
To benefit from labelled data, it has recently been proposed to consider a discriminative penalty term
along with the reconstruction error term in the optimisation process~\cite{Mairal_PAMI12}.
We are currently pursuing this line of research and seeking solutions for discriminative dictionary learning on Grassmann manifolds.
Moreover, the Frobenius norm used in our work is a special type of matrix Bregman divergence. Studying more involved cost functions derived from Bregman divergences is an interesting avenue to explore. On a side note, Bregman divergences induced from logdet function (\eg, Burg or symmetrical ones like Stein) cannot be directly used here because $\Mat{X}\Mat{X}^T$ is rank deficient.

\section*{Acknowledgements}
\label{sec:Acknowledgements}

NICTA is funded by the Australian Government
as represented by the {\it Department of Broadband, Communications and the Digital Economy},
as well as the Australian Research Council through the {\it ICT Centre of Excellence} program.
This work is funded in part through an ARC Discovery grant DP130104567. C. Shen's participation
was in part supported by ARC Future Fellowship F120100969.

\section*{Appendix}
\label{app:curve_equivalency}

Here, we prove Theorem \ref{thm:intrinsic_metric} (from Section~\ref{sec:proposed_method}),
\ie, the length of any given curve is the same under $\delta_g$ and $\delta_s$ up to a scale of $\sqrt{2}$.
\begin{proof}
Without any assumption on differentiability, let $(M,d)$ be a metric space.
A curve in $M$ is a continuous function $\gamma : [0, 1] \rightarrow M$ and joins the starting point $\gamma(0) = x$
to the end point $\gamma(1) = y$.
The intrinsic metric $\widehat{\delta}$	is defined as the infimum of the lengths of all paths from $x$ to $y$.
If the intrinsic metrics induced by two metrics $d_1$ and $d_2$ are identical to scale $\xi$,
then the length of any given curve is the same under both metrics up to $\xi$~\cite{Hartley_IJCV_13}.
\begin{theorem}
	If $d_1(x,y)$ and $d_2(x,y)$ are two metrics defined on a metric space $M$ such that
\begin{equation}
\lim_{d_1(x,y) \rightarrow 0} \:\frac{d_2(x,y)}{d_1(x,y)} = 1.
\label{eqn:intrinsic_metric0}
\end{equation}
uniformly (with respect to $x$ and $y$), then their intrinsic metrics are identical~\cite{Hartley_IJCV_13}.
\end{theorem}
Therefore, we need to study the behaviour of
\begin{equation*}
\lim_{\delta_g(\Mat{X},\Mat{Y}) \rightarrow 0} \:\frac{\delta_s(\Mat{X},\Mat{Y})}{\delta_g(\Mat{X},\Mat{Y})}
\end{equation*}
to prove our theorem on curve lengths.
We note that $\delta_s^2(\Mat{X},\Mat{Y}) = 2\sum_{i=1}^{p}\sin^2\theta_i$.
Since $\sin\theta_i \rightarrow \theta_i$ for $\theta_i \rightarrow 0$, we have
\begin{eqnarray*}
\lim_{\delta_g(\Mat{X},\Mat{Y}) \rightarrow 0} \:\frac{\delta_s^2(\Mat{X},\Mat{Y})}{\delta_g^2(\Mat{X},\Mat{Y})} =
\lim_{\delta_g(\Mat{X},\Mat{Y}) \rightarrow 0} \:\frac{2\sum_{i=1}^{p}\sin^2\theta_i}{\sum_{i=1}^{p}\theta_i^2}=
2\;,
\end{eqnarray*}%
\end{proof}

{{
\bibliographystyle{ieee}
\bibliography{references}
}
}

\end{document}